\documentclass[journal=jacsat,manuscript=article]{achemso}
\usepackage[version=3]{mhchem} 
\usepackage{siunitx}
\usepackage[dvipsnames]{xcolor}
\usepackage{amssymb}
\usepackage{pifont}
\usepackage{wrapfig}
\usepackage{hyperref}

\usepackage{amsmath} 

\usepackage{tabularx,colortbl}
\usepackage{capt-of}
\usepackage{adjustbox}

\author{Parisa Mollaei}
\affiliation[meche]
{Department of Mechanical Engineering, Carnegie Mellon University, 15213, USA}

\author{Amir Barati Farimani}
\email{barati@cmu.edu}
\affiliation[meche]
{Department of Mechanical Engineering, Carnegie Mellon University, 15213, USA}
\alsoaffiliation[biomed]
{Department of Biomedical Engineering, Carnegie Mellon University, 15213, USA}
\alsoaffiliation[mld]
{Machine Learning Department, Carnegie Mellon University, 15213, USA}

\title[An \textsf{achemso} demo]
{Protein Structure-Function Relationship: A Kernel-PCA Approach for Reaction Coordinate Identification}

\abbreviations{IR,NMR,UV}

\begin{document}

\begin{abstract}
In this study, we propose a Kernel-PCA model designed to capture structure-function relationships in a protein. This model also enables ranking of reaction coordinates according to their impact on protein properties. By leveraging machine learning techniques, including Kernel and principal component analysis (PCA), our model uncovers meaningful patterns in high-dimensional protein data obtained from molecular dynamics (MD) simulations. The effectiveness of our model in accurately identifying reaction coordinates has been demonstrated through its application to a G protein-coupled receptor. Furthermore, this model utilizes a network-based approach to uncover correlations in the dynamic behavior of residues associated with a specific protein property. These findings underscore the potential of our model as a powerful tool for protein structure-function analysis and visualization.
\end{abstract}

\clearpage

\section{Introduction}

Proteins serve as fundamental components of living organisms\cite{whitford2013proteins, simon1977organization, chothia1984principles}. The three-dimensional atomic arrangement determines structure, which in turn dictates their function. They are dynamic entities that undergo conformational changes in response to various stimuli or interactions with other molecules. It often leads to biological functions such as substance release or enzymatic activity.\cite{keskin2008principles, tompa2016principle, stollar2020uncovering}. Since structural changes directly influence biological function, deciphering structure-function relationships in proteins has long been a central pursuit in molecular biology. This endeavor is crucial for understanding the molecular mechanisms behind essential biological functions and diseases. 

To achieve this understanding, extensive structural data is required. Experimental biology methods typically provide limited structural data for proteins, usually only their stable conformations, such as folded and unfolded states. On the other hand, molecular dynamics (MD) simulations \cite{hollingsworth2018molecular, hospital2015molecular} have the capability to generate thousands of diverse protein structures, covering all possible states.
Determining structure-function relationships using MD simulation trajectories is also challenging due to their high dimensionality in space and sequential time dependence. A promising solution to this challenge is the efficient data reduction achieved through optimal protein representation. Protein representation refers to the way proteins are modeled and described in terms of their critical features.

This representation is capable of identifying functional sites, interaction networks, dynamic behavior, and conformational changes in biological processes. In the field of drug discovery\cite{david2020molecular, xiong2019pushing, lavecchia2015machine}, precise protein representation facilitates the discovery of potential drug targets and predicting the outcomes of molecular interactions. The representation is also vital in protein engineering\cite{alley2019unified, kouba2023machine, freschlin2022machine}, where the design of novel proteins with desired functions relies on a deep understanding of their structure-function relationships. In addition, the representation will aid in modeling and simulation of complex protein systems\cite{kolinski2004protein}. Such needs underscore the importance of developing a representation that captures both static and dynamic features determining protein properties. Moreover, it should effectively present the maximum variations in protein dynamics to minimize the data required for exploring structure-function relationships. In this study, we introduce a model aimed at creating such a representation while addressing critical questions, including the following: Is this model generalizable to all proteins with different sequences and structures, when utilizing raw atomic coordinates? Is it feasible to use only a subset of atoms to have a reduced representation? How could we reveal the reaction coordinates using this representation? Is the model capable of identifying the extent of contribution of individual residues to the overall protein properties?. To answer these questions, we took advantage of machine learning (ML) methods, as they offer computational tools and methodologies essential to dealing with high-dimensional protein data\cite{tarca2007machine, majaj2018deep, yadav2022prediction, guntuboina2023peptidebert, mollaei2023activity, kim2024gpcr, baldi2001bioinformatics, mollaei2023unveiling, mollaei2024idp, camacho2018next, badrinarayanan2024multi}. 

The two well-known techniques for feature extraction and dimensionality reduction are Kernels\cite{kung2014kernel, gehler2009introduction} and Principal Component Analysis (PCA)\cite{abdi2010principal, mackiewicz1993principal}, respectively. Kernels offer a flexible framework for projecting datasets into higher-dimensional spaces. This enables the transformation of protein data into feature map spaces that enhance learning. PCA is a widely used method to reduce the dimensionality of data while preserving its essential features. It is a linear technique that captures the directions (principle components) in which the data vary the most. By integrating these two methods, we developed a model that effectively analyzes protein structure-function relationships using MD trajectories and provides meaningful representation for it.

 \begin{figure}[t!]
     \centering
     \includegraphics[width=\linewidth]{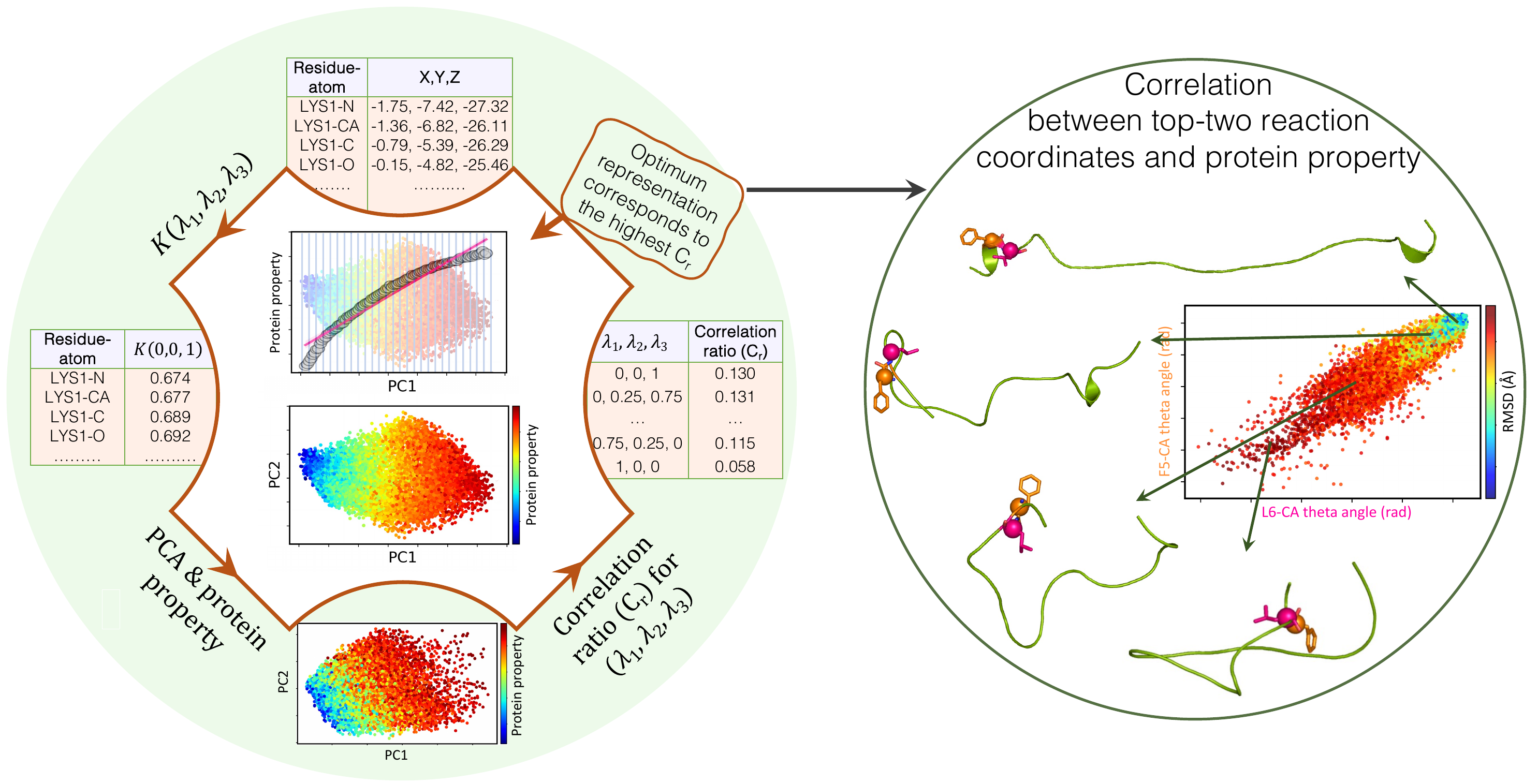}
     \caption{Overview of Kernel-PCA model. Trajectories are initially prepared to provide atomic coordinates as input for a Kernel model (K($\lambda_1$, $\lambda_2$, $\lambda_3$)). It is followed by PCA to generate 2D representations for Kernel outputs. Finally, a defined Correlation ratio ($C_r$) selects the optimal representation. This representation identifies and ranks reaction coordinates based on their significance to the protein property while also uncovering correlations among them. }
     \label{fig:framework}
 \end{figure}

\section{Method}

Our analysis begins with the preparation of the protein trajectories, as outlined in detail in Supporting Information Section 1. Figure.\ref{fig:framework} illustrates the framework where raw atomic coordinates serve as input for a Kernel model (Section I). The Kernel outputs are then processed by PCA\cite{abdi2010principal, mackiewicz1993principal} to extract the essential features. Later, the optimum representation is selected via a defined Correlation ratio (Section II). Ultimately, this representation introduces reaction coordinates and highlights their extent of correlations with protein properties in sequential order (Section III).

\subsection*{I. Kernel Design for Effective Representation}

\begin{figure}[t!]
     \centering
     \includegraphics[width=\linewidth]{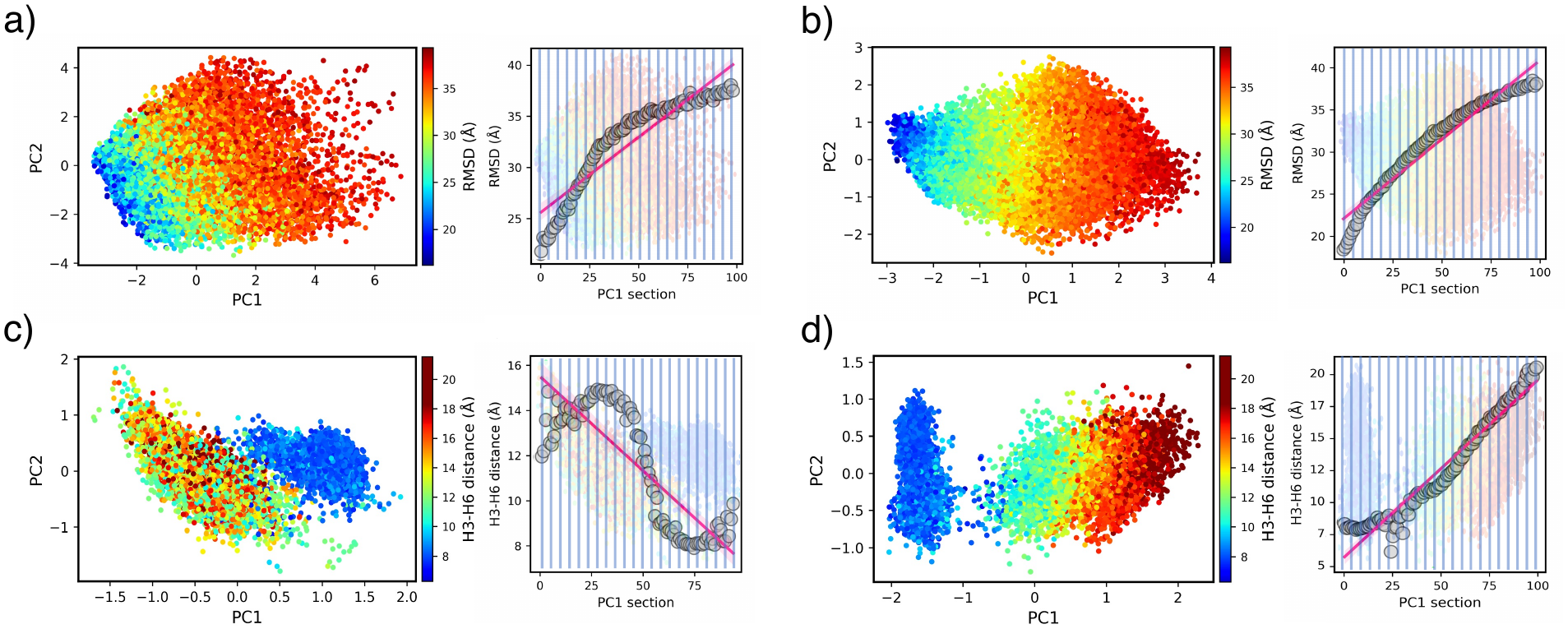}
     \caption{Kernel-PCA representations of the NTL9 protein (a, b) and $\beta_2$ adrenergic receptor (c, d) corresponding to the lowest (a, c) and highest (b, d) $C_r$ values. For the NTL9 protein: (a) K(0, 1, 0) with $C_r$ = 0.05, (b) K(0.5,0.5,0) with $C_r$ = 0.15. For the $\beta_2$ adrenergic receptor c) K(0,1,0) with $C_r$ = 0.03, d) K(0.75,0,0.25) with $C_r$ = 0.13.}
     \label{fig:cr}
 \end{figure}

Why is a Kernel model necessary? Initially, we opted to use PCA \cite{abdi2010principal, mackiewicz1993principal} on raw data due to its ability to effectively identify principal components. These components are linear combinations of original features achieved by projecting high-dimensional data onto a 2D subspace. This subspace allows efficient visualization and analysis. However, PCA alone may face challenges in extracting the intricate relationships within atom dynamics and their correlations to overall protein properties when utilizing the raw coordination of atoms. This can be attributed to several factors, including the high dimensionality of the data, which can lead to computational costs and overfitting. Moreover, some raw coordinate data may have less crucial structural information, making it challenging for PCA to obtain maximum variations effectively. Additionally, the presence of potential noises in the raw coordinates can further hinder the performance of PCA. Furthermore, PCA is a linear model, whereas complex coordinate data may exhibit nonlinear relationships. These challenges become particularly amplified when our goal is to generate a representation in which PC1 predominantly captures reaction coordinates and exhibits a strong relationship (roughly linear) with protein properties (Figure \ref{fig:cr} b,d). This choice of representation is motivated by its ability to significantly improve the identification of reaction coordinates. To address these challenges, we initially developed a Kernel model\cite{kung2014kernel, gehler2009introduction}. 

In our previous work\cite{mollaei2023unveiling}, we demonstrated that the angles formed between atoms within individual residues are particularly valuable for identifying switch residues, which exhibit transitions between angular states correlated with protein function. 
Building on this foundation, we developed an angular Kernel model that utilizes atomic coordinates to capture conformational changes in proteins. We aimed to transform these data into a feature map space to uncover conformation-function relationships. Later, the PCA effectively reduces dimensionality of the transformed space to 2D while retaining crucial information. The Kernel model is defined as follows:

\vspace{-\baselineskip}
\begin{center}
\[
Kernel = \lambda_{1} \left( \cos\left(\frac{x}{r}\right) \right)^2 + \lambda_{2}\left( \cos\left(\frac{y}{r}\right) \right)^2 + \lambda_{3} \left( \sin\left(\frac{z}{r}\right) \right)^2
\]
\end{center}

\noindent Where (x, y, z) represent the coordinates of individual atoms and ($\lambda_1$, $\lambda_2$, $\lambda_3$) are hyperparameters. We established two constraints to limit the possibilities for hyperparameters and narrow the search space. We ensured that each $\lambda_i$ fell within the range [0, 1] and sum of coefficients equaled to 1. In this study, we selected five values for $\lambda_i$ as [0, 0.25, 0.5, 0.75, 1], resulting in 15 combinations for ($\lambda_1$, $\lambda_2$, $\lambda_3$) that confirm the summation equals 1. 

\noindent It is important to note that studies have shown that the Kernel model enhances PCA performance compared to relying solely on x, y, or z coordinates (Supporting Information Section 2). Figure.\ref{fig:cr} illustrates the impact of $\lambda_i$ values in representations of NTL9 protein and $\beta_2$ adrenergic receptor. For both cases, K(0,1,0) generates representations where neither PC1 nor PC2 exhibit specific correlation with the proteins properties (Figure.\ref{fig:cr}a,c), whereas this correlation is notably stronger with PC1 alone for K(0.5,0.5,0) and K(0.75,0,0.25) in the NTL9 protein and $\beta_2$ adrenergic receptor, respectively (Figure.\ref{fig:cr}b,d). After determining the optimal representation, we will elaborate on how to identify reaction coordinates using this correlation between protein properties and PC1.

\subsection{II. Evaluation of Representations Using Correlation Ratio}

Among different combinations of $\lambda_{i}$ values, one of them yields a representation that provides the highest correlation between PC1 and protein properties. To automate the identification of that one, we have established a metric for evaluating the Kernel's performance based on the correlation between PC1 and protein property. We first divide the PC1 space into equal sections and calculate the average value of protein property in each section. We then fit the best linear curve on the average values to find its slope ($S$) and the coefficient of determination ($R^2$). In addition, we measured average of total variances of the protein property within individual sections ($V$). The Correlation ratio ($C_r$) is defined as follows:

\vspace{-\baselineskip}
\begin{center}
\[ \text{$C_r$} = \frac{{\text{{$S$}} \times \text{{$R^2$}}}}{{\text{{$V$}}^{0.5}}}
\]
\end{center}

\noindent Where the highest $C_r$ introduces the optimal representation with the strongest correlation between protein properties and PC1. We assessed the impact of fifteen combinations of $\lambda_1$, $\lambda_2$, and $\lambda_3$ on Correlation ratios for Protein B, NTL9, Trp-Cage, and Chignolin. For details, see Supporting Information Section 3.

\noindent Generating the optimal representation involved using all atomic coordinates within each structure across thousands of trajectories. However, this approach will lead to high computational costs for large proteins. We experimented with using only one single type of atom to develop a reduced representation. Initially, we considered individual atom types (such as $O$, $N$, $C$, $CA$, etc.) and used each one's coordinates for the Kernel-PCA model. Ultimately, $CB$ atoms exhibit the highest $C_r$ value compared to other atoms. We observed that in large proteins like $\beta_2$ adrenergic receptor, using only $CB$ atoms yields a higher $C_r$ value compared to using all atom types. This improvement may stem from filtering out noise, specifically less important atomic information, from the input data.

To evaluate the effectiveness of $CB$ representation versus all-atoms representation, we defined $R_{CB,T}$ as the ratio of $C_r$ for only $CB$ atoms to $C_r$ for total atoms. $R_{CB,T}$ value for $\beta_2$ adrenergic receptor is 111.27\%. For Protein B, NTL9, Trp-Cage, and Chignolin $R_{CB,T}$ values are 99.64\%, 89.23\% ,93.01\% and 87.84\%, respectively. $R_{CB,T}$ values indicate that both using all atoms and using only $CB$ atoms yield mostly similar results for representation as well as reaction coordinates. This approach led to a significant reduction in the feature space. For example, in the case of the $\beta_2$ adrenergic receptor, it decreased from 2175 (total atoms) to 260 ($CB$ atoms) in a single structure. This reduction simplifies the model complexity, improves computational efficiency, and focuses on more relevant features, potentially enhancing Kernel-PCA model performance.

\subsection{III. Reaction Coordinate Identification}

The aim of generating the optimum representation is to obtain the ranked reaction coordinates. These coordinates serve as essential variables that distill the complex protein system into essential variables that govern the protein's function and biochemical processes. In this study, the ranked reaction coordinates refer to the structural features contributing to the protein's properties arranged according to their level of association. For the reaction coordinate analysis, we focused on dynamics of individual amino acids in proteins. Since the dynamic of $\alpha$-carbon atom ($CA$) is sufficient to learn dynamic of entire amino acid\cite{doerr2021torchmd}, we measured theta angles ($\arccos(z/r)$) of the atom in every single residue over proteins. Each residue's theta angles were then projected onto the optimum representation for $C_r$ calculation. Since the highest $C_r$ value for the protein property identified the optimal representation, the $C_r$ values of individual features over this representation effectively measure their contribution to the property. As a result, we ranked the theta angles (reaction coordinates) based on their $C_r$ values. The top-ranked ones reflect the most dominant modes of motion at residue level driving the protein's property.

\section{Dataset and MD Simulation}

To assess the effectiveness and generalizability of our model, we benchmarked it using one large protein and four small proteins. For the large one, we selected the $\beta_2$ adrenergic receptor, which consists of seven transmembrane helices (PDB: 2RH1). The original MD simulation dataset\cite{kohlhoff2014cloud} was performed with the partial inverse agonist carazolol. The receptor is embedded within a POPC lipid bilayer and solvated with TIP3P water molecules\cite{jorgensen1983comparison}. The simulations were executed in parallel on Google’s Exacycle platform\cite{hellerstein2012science} and performed using the Gromacs 4.5.3 MD package\cite{hess2008gromacs}. Using these simulations, we can explore the activation process correlated with conformational changes in the receptor. 
For this protein, we randomly sampled 10,000 structures within 5000 MD simulations.

\noindent For the small proteins, we used all-atom MD simulation trajectories for Protein B (PDB: 1PRB), NTL9 (PDB: 2HBA), Trp-Cage (PDB: 2JOF), and Chignolin (PDB: 5AWL)\cite{majewski2023machine, kubelka2004protein, lindorff2011fast}. These simulations investigated the folding process in these proteins with diverse structures, ranging in length from 10 to 47 amino acids. Protein B and Trp-Cage contain $\alpha$ helices, while Chignolin stands out as a purely $\beta$-sheet protein and NTL9 displays mixed $\alpha$$\beta$ structures. To generate the original dataset, MD simulations were initiated from random coil conformations for each protein. They involved solvating all initial structures in cubic boxes, ionizing them according to Lindorff-Larsen et al\cite{lindorff2011fast} protocol, and employing ACEMD\cite{harvey2009acemd} on the GPUGRID.net network\cite{buch2010high}. The CHARMM22*\cite{piana2011robust} force field and TIP3P water model\cite{jorgensen1983comparison} were utilized at a temperature of 350K. For these proteins, we randomly sampled 10,000 structures from the original MD simulation trajectories. 

\section{Results}

We aim to find optimum representations and capture the top-ranked reaction coordinates. The diverse set of proteins, varying in size and structure with different properties allows us to assess the generalizability of our model. 

\subsection{Case 1: Reaction Coordinates in $\beta_2$ Adrenergic Receptor}

\begin{figure}[t!]
 \centering
 \includegraphics[width=\linewidth]{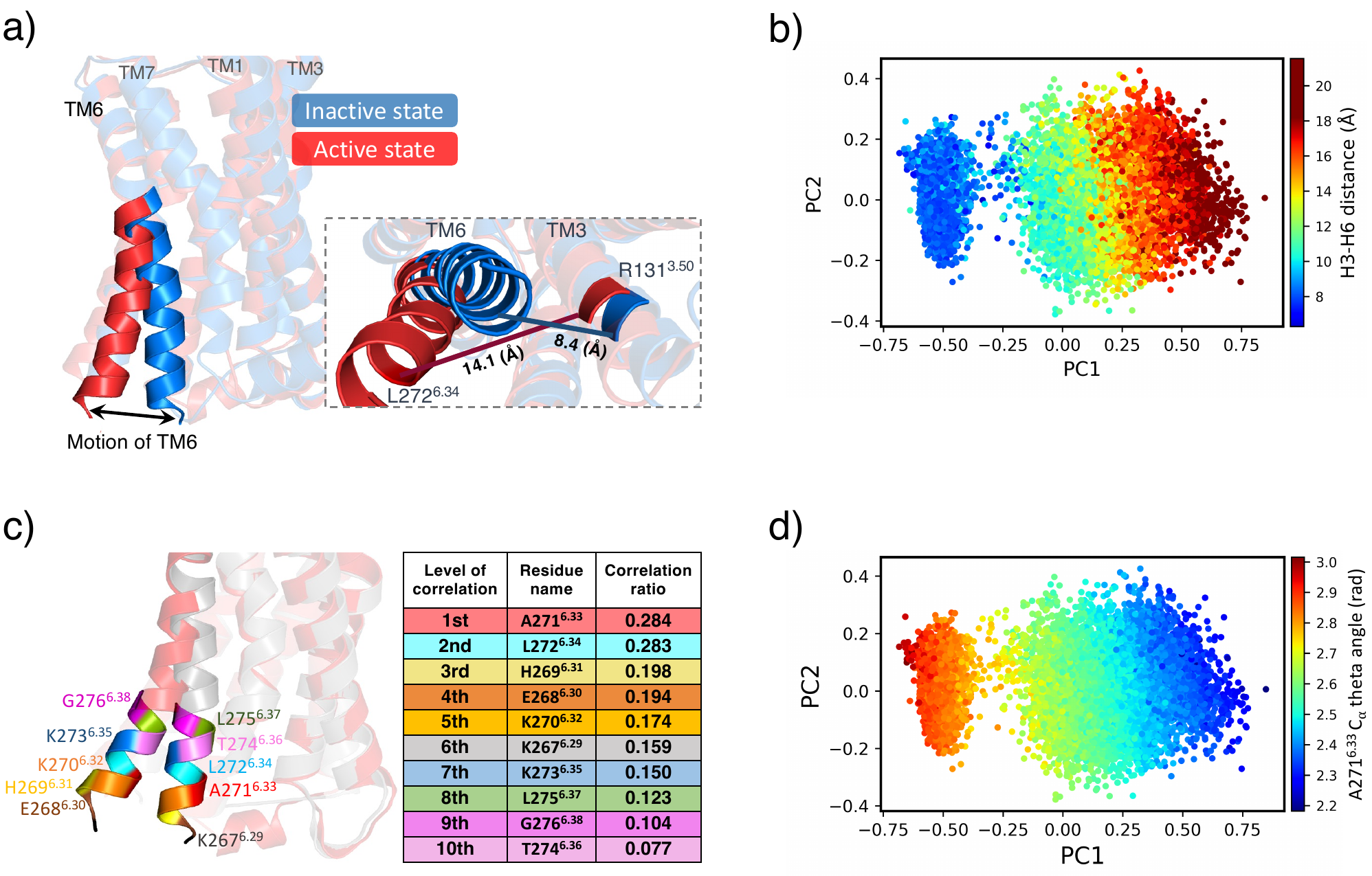}
 \caption{Identification of reaction coordinates in $\beta_2AR$ using Kernel-PCA model. a) The known reaction coordinates in $\beta_2AR$ are the intracellular parts of TM6 during the transition from inactive to active state b) H3-H6 distances mapped onto the representation. c) amino acid types located in where essential conformational changes occur during activation process. d) top ten PC1 reaction coordinates, indicating dynamics of residues with the highest contribution to activation states of the receptor. e) dynamic motion of ${A271}^{6.33}$ projected onto the PC representation.}
 \label{fig:tm6}
\end{figure}

G protein-coupled receptors (GPCRs) are a diverse group of cell surface receptors that undergo conformational changes while interacting with ligands\cite{rosenbaum2009structure, katritch2013structure, schoneberg1999structural}. Due to ligand binding, GPCRs become activated and trigger intracellular signaling cascades, leading to various physiological processes within the body\cite{wang2018new, heng2013overview, hilger2018structure}. GPCRs consist of seven transmembrane helices (TMs), among which the key conformational change during receptor activation primarily involves the movement of transmembrane helix six (TM6)\cite{simpson2011modeling, latorraca2017gpcr}. We analyzed structure-function relationship in $\beta_2$ adrenergic receptor ($\beta_2AR$)\cite{ali2020beta, yang2021role}. Figure.\ref{fig:tm6}a depicts the motion of TM6 between the inactive (blue: PDB 2RH1)\cite{cherezov2007high} and active (red: PDB 3P0G)\cite{rasmussen2011structure} structures of $\beta_2AR$. Typically, this conformational change is identified by measuring the contact distance between TM6 and TM3\cite{kohlhoff2014cloud}. In this study, we measured the C$_\alpha$ contact distance between ${R131}^{3.50}$-${L272}^{6.34}$ residues in the receptor (denoted H3-H6 distance). Figure.\ref{fig:tm6}a also highlights H3-H6 distance increases from 8.4 $\AA$ to 14.1 $\AA$ in the transitions from inactive to active states. Given the size of this receptor, we utilized only the $CB$ atoms when generating representations (see Section II). Due to the strong correlation between H3-H6 distances and activation states, we established the optimal representation based on these distances. Figure.\ref{fig:tm6}b illustrates that the chosen representation offers a strong correlation of receptor activation with PC1. Therefore, we can analyze reaction coordinates within this representation, as discussed in Section III.
Figure \ref{fig:tm6}c displays the top ten reaction coordinates in the receptor's conformation that are marked with colors, correspond to those listed in the table. Notably, all of them are located at the intracellular part of TM6, where the receptor undergoes the most significant motion during the activation process. This attests to the strength of our model in capturing the essential structural features correlated to the protein's property. 

Figure.\ref{fig:tm6}d demonstrates the dynamic motion of ${A271}^{6.33}$, as the first-ranked reaction coordinate, mapped on the selected representation. Comparing it with Figure.\ref{fig:tm6}b, we observe the theta angle in ${A271}^{6.33}$ residue decreases when the protein becomes activated. Such relationships can be identified by comparing pattern of distribution of each feature across the representation with its corresponding pattern in the property. For further investigation, the representations of reaction coordinates listed in Figure.\ref{fig:tm6}c are illustrated in Supporting Information Section 4, Figure.S4.

\subsection{Case 2: Reaction Coordinates in Protein B, NTL9, Trp-Cage, and Chignolin }

We assessed the folding states of Protein B (PDB: 1PRB), NTL9 (PDB: 2HBA), Trp-Cage (PDB: 2JOF), and Chignolin (PDB: 5AWL) proteins by measuring the Root Mean Squared Displacement (RMSD)\cite{mcgibbon2015mdtraj} relative to their non-folded reference structures (see Supporting Information Section 1). For these proteins, Figure.S4 displays normalized $C_r$ values plotted against $\lambda_i$ values, illustrating how different combinations of $\lambda_i$ affect the $C_r$ value. Moreover, Table.S1 lists the $\lambda_i$ values corresponding to the highest and lowest $C_r$ values.

\begin{figure}[t!]
     \centering
    \includegraphics[width=\linewidth]{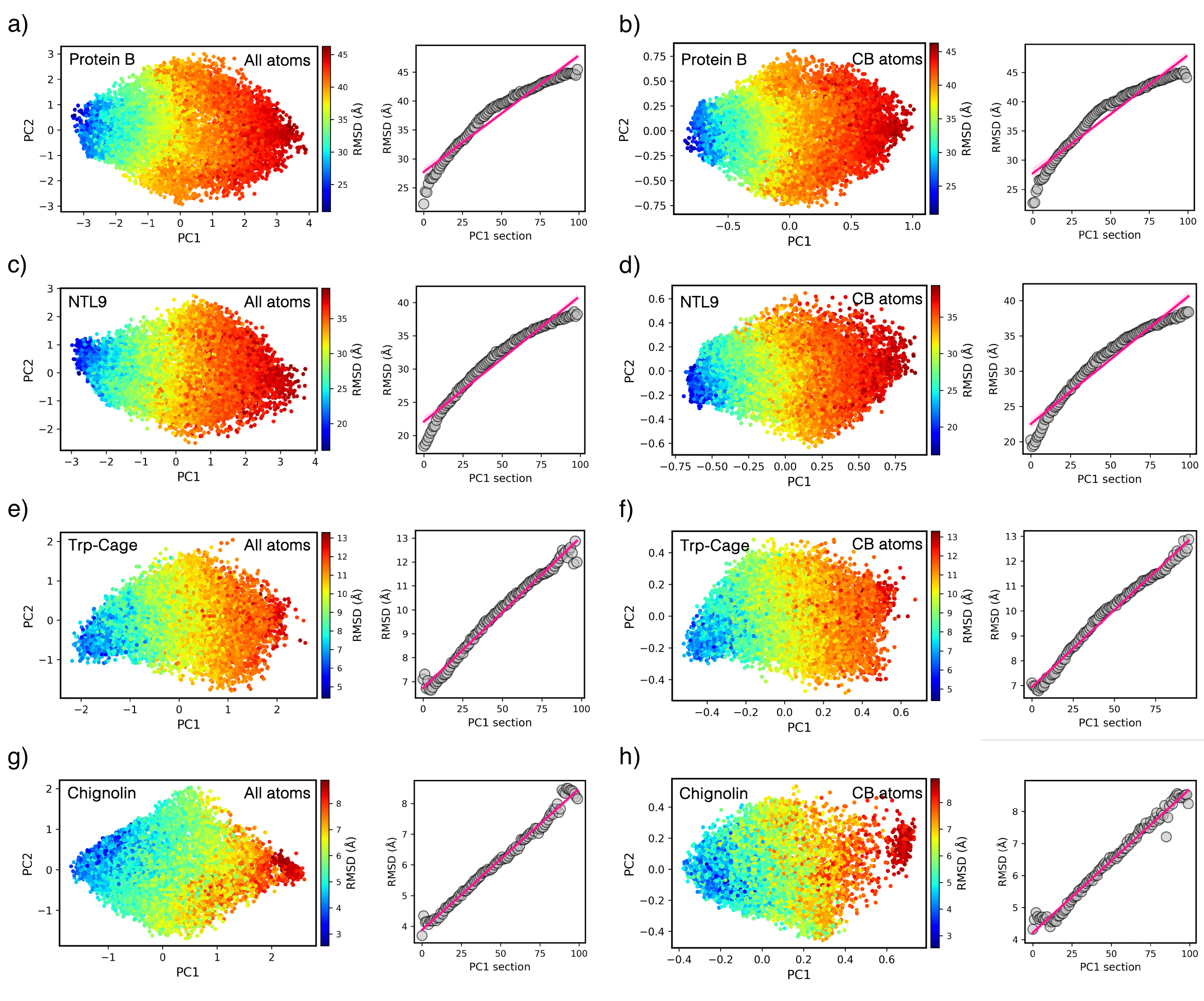}
     \caption{RMSDs projected onto the optimal representation, illustrating the maximum Correlation ratio achieved using all atoms for (a) Protein B, (c) NTL9, (e) Trp-Cage, and (g) Chignolin. Panels (b), (d), (f), and (h) show their corresponding representations generated using only the $CB$ atoms.}
     \label{fig:all}
 \end{figure}

For the small proteins, Figure.\ref{fig:all} depicts the optimum representations generated using all atoms as well as only the $CB$ atoms, colored by RMSDs values. The figures are followed by corresponding fitted lines that maximize $C_r$ values. Table 1 lists the top three reaction coordinates in each protein, introducing the key residues correlated with folding states.

\begin{table}[t!]
\caption{The top three most significant reaction coordinates in Protein B, NTL9, Trp-Cage, and Chignolin, introducing the motion dynamics of residues with the highest contribution to the overall RMSD.} 
\label{tab:my table} 
\includegraphics[width=0.55\linewidth]{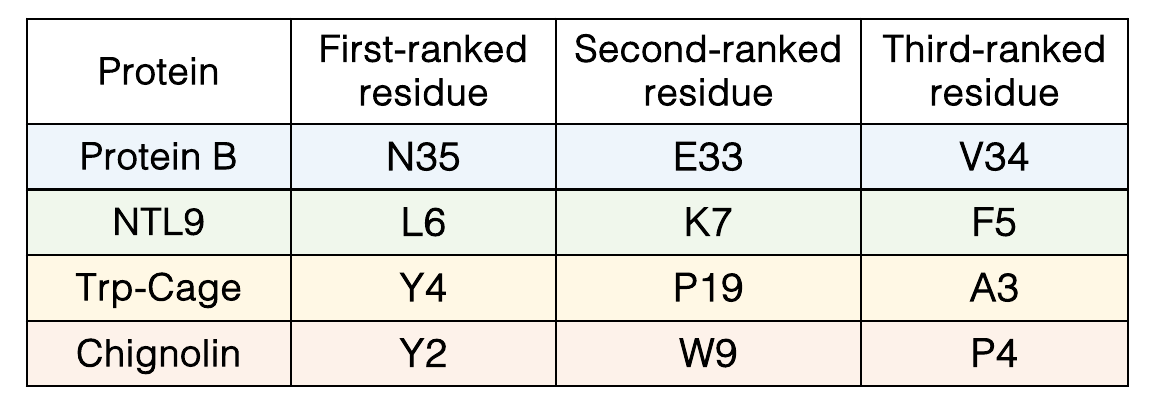}
\end{table}
 
\noindent  It is obvious that even though the property is common among four proteins, the top-ranked residues are recognized in different regions of each protein, which offers the possibility that the folding mechanism may vary across these proteins. More studies are required to understand the role of N35, L6, Y4, and Y2 as the first-ranked reaction coordinates in Protein B, NTL9, Trp-Cage, and Chignolin, respectively, in their folding mechanisms. Extending this list to include all residues as the ranked reaction coordinates allows us to analyze the internal interactions of each residue, which demonstrate their collective impact in harmony.

\section{Network-Based Insights into Reaction Coordinate Relationships Linked to Protein Property}
\begin{figure}[t!]
     \centering
     \includegraphics[width=0.96\linewidth]{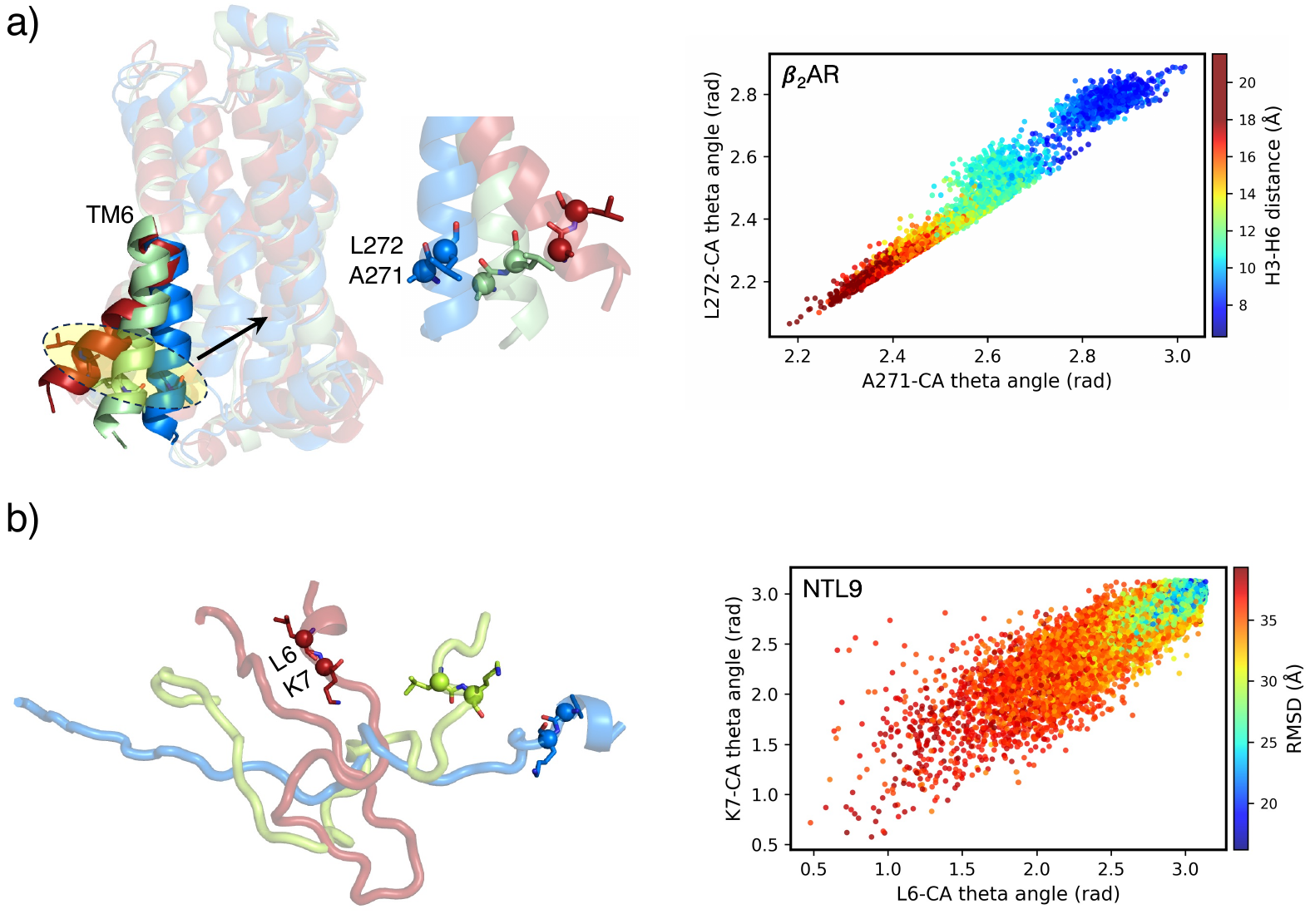}
     \caption{Strong correlation between the top two residues associated with the protein property. a) relationship between ${A271}^{6.33}$ and ${L272}^{6.34}$ residues in the inactive (blue), intermediate (green), and active (red) states of $\beta_2AR$. b) interaction between L6 and K7 residues in the unfolded (blue), intermediate (green), and folded (red) states in NTL9 protein.}
     \label{fig:2topatoms}
 \end{figure}

We delved deeper to establish connections among dynamical behaviors of reaction coordinates linked to protein property (Figure \ref{fig:2topatoms}). Our findings reveal that pairs of top-ranked reaction coordinates form a harmonious linear and diagonal pattern when strongly associated with protein property. For example, in $\beta_2AR$, theta angles in residues ${A271}^{6.33}$ and ${L272}^{6.34}$ decrease simultaneously as the receptor becomes activated (Figure \ref{fig:2topatoms}a). A similar pattern is observed in the L6 and K7 residues in NTL9 protein during the folding process, where both angles decrease as the protein folds. See Supporting Information Section 5 for correlation between the top-ranked reaction coordinates in Protein B, Trp-Cage, and Chignolin proteins. Exploring such relationships across all residues provides insights into how a particular interaction influences protein dynamics and function. This network-based approach can guide targeted drug design and therapeutic interventions by identifying crucial interaction sites and their impact on protein behavior.

\section{Conclusion}

In this study, we introduced a Kernel-PCA model for efficient protein representation, aiming to elucidate structure-function relationships by identifying reaction coordinates. These reaction coordinates are then ranked on the basis of their degree of contribution to the protein property. The flexibility of the representation, along with tunable hyperparameters, ensures applicability to various protein dynamics obtained through MD simulations. We initially evaluated the efficacy of our model by applying it to the $\beta_2$ adrenergic receptor, a G protein-coupled receptor with a known structure-activation relationship. Subsequently, we applied the model to Protein B, NTL9, Trp-Cage, and Chignolin with unknown relationships between structural features and folding processes. Using this model, we can also establish interconnections among the dynamics of all residues linked to the protein property through a network-based approach. This method provides a robust tool for understanding complex biological systems and advancing biotechnology applications.


\section{Data and software availability}
The necessary information containing the codes and data for downstream tasks used in this study is available here: https://github.com/pmollaei/ProtKernel

\section{Supporting Information}
The Supporting Information includes data preprocessing, protein representation using atomic coordinates, Correlation ratios correspond to K($\lambda_1$, $\lambda_2$, $\lambda_3$), top reaction coordinates in $\beta_2$ Adrenergic receptor, and relationships between top reaction coordinates in Protein B, Trp-Cage, and Chignolin proteins.

\begin{acknowledgement}

This work is supported by the Center for Machine Learning in Health (CMLH) at Carnegie Mellon University and a start-up fund from the Mechanical Engineering Department at CMU. 

\end{acknowledgement}

\bibliography{reference}

\providecommand{\latin}[1]{#1}
\makeatletter
\providecommand{\doi}
  {\begingroup\let\do\@makeother\dospecials
  \catcode`\{=1 \catcode`\}=2 \doi@aux}
\providecommand{\doi@aux}[1]{\endgroup\texttt{#1}}
\makeatother
\providecommand*\mcitethebibliography{\thebibliography}
\csname @ifundefined\endcsname{endmcitethebibliography}  {\let\endmcitethebibliography\endthebibliography}{}
\begin{mcitethebibliography}{55}
\providecommand*\natexlab[1]{#1}
\providecommand*\mciteSetBstSublistMode[1]{}
\providecommand*\mciteSetBstMaxWidthForm[2]{}
\providecommand*\mciteBstWouldAddEndPuncttrue
  {\def\EndOfBibitem{\unskip.}}
\providecommand*\mciteBstWouldAddEndPunctfalse
  {\let\EndOfBibitem\relax}
\providecommand*\mciteSetBstMidEndSepPunct[3]{}
\providecommand*\mciteSetBstSublistLabelBeginEnd[3]{}
\providecommand*\EndOfBibitem{}
\mciteSetBstSublistMode{f}
\mciteSetBstMaxWidthForm{subitem}{(\alph{mcitesubitemcount})}
\mciteSetBstSublistLabelBeginEnd
  {\mcitemaxwidthsubitemform\space}
  {\relax}
  {\relax}

\bibitem[Whitford(2013)]{whitford2013proteins}
Whitford,~D. \emph{Proteins: structure and function}; John Wiley \& Sons, 2013\relax
\mciteBstWouldAddEndPuncttrue
\mciteSetBstMidEndSepPunct{\mcitedefaultmidpunct}
{\mcitedefaultendpunct}{\mcitedefaultseppunct}\relax
\EndOfBibitem
\bibitem[Simon and Simon(1977)Simon, and Simon]{simon1977organization}
Simon,~H.~A.; Simon,~H.~A. The organization of complex systems. \emph{Models of discovery: And other topics in the methods of science} \textbf{1977}, 245--261\relax
\mciteBstWouldAddEndPuncttrue
\mciteSetBstMidEndSepPunct{\mcitedefaultmidpunct}
{\mcitedefaultendpunct}{\mcitedefaultseppunct}\relax
\EndOfBibitem
\bibitem[Chothia(1984)]{chothia1984principles}
Chothia,~C. Principles that determine the structure of proteins. \emph{Annual review of biochemistry} \textbf{1984}, \emph{53}, 537--572\relax
\mciteBstWouldAddEndPuncttrue
\mciteSetBstMidEndSepPunct{\mcitedefaultmidpunct}
{\mcitedefaultendpunct}{\mcitedefaultseppunct}\relax
\EndOfBibitem
\bibitem[Keskin \latin{et~al.}(2008)Keskin, Gursoy, Ma, and Nussinov]{keskin2008principles}
Keskin,~O.; Gursoy,~A.; Ma,~B.; Nussinov,~R. Principles of protein- protein interactions: what are the preferred ways for proteins to interact? \emph{Chemical reviews} \textbf{2008}, \emph{108}, 1225--1244\relax
\mciteBstWouldAddEndPuncttrue
\mciteSetBstMidEndSepPunct{\mcitedefaultmidpunct}
{\mcitedefaultendpunct}{\mcitedefaultseppunct}\relax
\EndOfBibitem
\bibitem[Tompa(2016)]{tompa2016principle}
Tompa,~P. The principle of conformational signaling. \emph{Chemical Society Reviews} \textbf{2016}, \emph{45}, 4252--4284\relax
\mciteBstWouldAddEndPuncttrue
\mciteSetBstMidEndSepPunct{\mcitedefaultmidpunct}
{\mcitedefaultendpunct}{\mcitedefaultseppunct}\relax
\EndOfBibitem
\bibitem[Stollar and Smith(2020)Stollar, and Smith]{stollar2020uncovering}
Stollar,~E.~J.; Smith,~D.~P. Uncovering protein structure. \emph{Essays in biochemistry} \textbf{2020}, \emph{64}, 649--680\relax
\mciteBstWouldAddEndPuncttrue
\mciteSetBstMidEndSepPunct{\mcitedefaultmidpunct}
{\mcitedefaultendpunct}{\mcitedefaultseppunct}\relax
\EndOfBibitem
\bibitem[Hollingsworth and Dror(2018)Hollingsworth, and Dror]{hollingsworth2018molecular}
Hollingsworth,~S.~A.; Dror,~R.~O. Molecular dynamics simulation for all. \emph{Neuron} \textbf{2018}, \emph{99}, 1129--1143\relax
\mciteBstWouldAddEndPuncttrue
\mciteSetBstMidEndSepPunct{\mcitedefaultmidpunct}
{\mcitedefaultendpunct}{\mcitedefaultseppunct}\relax
\EndOfBibitem
\bibitem[Hospital \latin{et~al.}(2015)Hospital, Go{\~n}i, Orozco, and Gelp{\'\i}]{hospital2015molecular}
Hospital,~A.; Go{\~n}i,~J.~R.; Orozco,~M.; Gelp{\'\i},~J.~L. Molecular dynamics simulations: advances and applications. \emph{Advances and Applications in Bioinformatics and Chemistry} \textbf{2015}, 37--47\relax
\mciteBstWouldAddEndPuncttrue
\mciteSetBstMidEndSepPunct{\mcitedefaultmidpunct}
{\mcitedefaultendpunct}{\mcitedefaultseppunct}\relax
\EndOfBibitem
\bibitem[David \latin{et~al.}(2020)David, Thakkar, Mercado, and Engkvist]{david2020molecular}
David,~L.; Thakkar,~A.; Mercado,~R.; Engkvist,~O. Molecular representations in AI-driven drug discovery: a review and practical guide. \emph{Journal of Cheminformatics} \textbf{2020}, \emph{12}, 56\relax
\mciteBstWouldAddEndPuncttrue
\mciteSetBstMidEndSepPunct{\mcitedefaultmidpunct}
{\mcitedefaultendpunct}{\mcitedefaultseppunct}\relax
\EndOfBibitem
\bibitem[Xiong \latin{et~al.}(2019)Xiong, Wang, Liu, Zhong, Wan, Li, Li, Luo, Chen, Jiang, \latin{et~al.} others]{xiong2019pushing}
Xiong,~Z.; Wang,~D.; Liu,~X.; Zhong,~F.; Wan,~X.; Li,~X.; Li,~Z.; Luo,~X.; Chen,~K.; Jiang,~H.; others Pushing the boundaries of molecular representation for drug discovery with the graph attention mechanism. \emph{Journal of medicinal chemistry} \textbf{2019}, \emph{63}, 8749--8760\relax
\mciteBstWouldAddEndPuncttrue
\mciteSetBstMidEndSepPunct{\mcitedefaultmidpunct}
{\mcitedefaultendpunct}{\mcitedefaultseppunct}\relax
\EndOfBibitem
\bibitem[Lavecchia(2015)]{lavecchia2015machine}
Lavecchia,~A. Machine-learning approaches in drug discovery: methods and applications. \emph{Drug discovery today} \textbf{2015}, \emph{20}, 318--331\relax
\mciteBstWouldAddEndPuncttrue
\mciteSetBstMidEndSepPunct{\mcitedefaultmidpunct}
{\mcitedefaultendpunct}{\mcitedefaultseppunct}\relax
\EndOfBibitem
\bibitem[Alley \latin{et~al.}(2019)Alley, Khimulya, Biswas, AlQuraishi, and Church]{alley2019unified}
Alley,~E.~C.; Khimulya,~G.; Biswas,~S.; AlQuraishi,~M.; Church,~G.~M. Unified rational protein engineering with sequence-based deep representation learning. \emph{Nature methods} \textbf{2019}, \emph{16}, 1315--1322\relax
\mciteBstWouldAddEndPuncttrue
\mciteSetBstMidEndSepPunct{\mcitedefaultmidpunct}
{\mcitedefaultendpunct}{\mcitedefaultseppunct}\relax
\EndOfBibitem
\bibitem[Kouba \latin{et~al.}(2023)Kouba, Kohout, Haddadi, Bushuiev, Samusevich, Sedlar, Damborsky, Pluskal, Sivic, and Mazurenko]{kouba2023machine}
Kouba,~P.; Kohout,~P.; Haddadi,~F.; Bushuiev,~A.; Samusevich,~R.; Sedlar,~J.; Damborsky,~J.; Pluskal,~T.; Sivic,~J.; Mazurenko,~S. Machine learning-guided protein engineering. \emph{ACS catalysis} \textbf{2023}, \emph{13}, 13863--13895\relax
\mciteBstWouldAddEndPuncttrue
\mciteSetBstMidEndSepPunct{\mcitedefaultmidpunct}
{\mcitedefaultendpunct}{\mcitedefaultseppunct}\relax
\EndOfBibitem
\bibitem[Freschlin \latin{et~al.}(2022)Freschlin, Fahlberg, and Romero]{freschlin2022machine}
Freschlin,~C.~R.; Fahlberg,~S.~A.; Romero,~P.~A. Machine learning to navigate fitness landscapes for protein engineering. \emph{Current opinion in biotechnology} \textbf{2022}, \emph{75}, 102713\relax
\mciteBstWouldAddEndPuncttrue
\mciteSetBstMidEndSepPunct{\mcitedefaultmidpunct}
{\mcitedefaultendpunct}{\mcitedefaultseppunct}\relax
\EndOfBibitem
\bibitem[Koli{\'n}ski \latin{et~al.}(2004)Koli{\'n}ski, \latin{et~al.} others]{kolinski2004protein}
Koli{\'n}ski,~A.; others Protein modeling and structure prediction with a reduced representation. \emph{Acta Biochimica Polonica} \textbf{2004}, \emph{51}\relax
\mciteBstWouldAddEndPuncttrue
\mciteSetBstMidEndSepPunct{\mcitedefaultmidpunct}
{\mcitedefaultendpunct}{\mcitedefaultseppunct}\relax
\EndOfBibitem
\bibitem[Tarca \latin{et~al.}(2007)Tarca, Carey, Chen, Romero, and Dr{\u{a}}ghici]{tarca2007machine}
Tarca,~A.~L.; Carey,~V.~J.; Chen,~X.-w.; Romero,~R.; Dr{\u{a}}ghici,~S. Machine learning and its applications to biology. \emph{PLoS computational biology} \textbf{2007}, \emph{3}, e116\relax
\mciteBstWouldAddEndPuncttrue
\mciteSetBstMidEndSepPunct{\mcitedefaultmidpunct}
{\mcitedefaultendpunct}{\mcitedefaultseppunct}\relax
\EndOfBibitem
\bibitem[Majaj and Pelli(2018)Majaj, and Pelli]{majaj2018deep}
Majaj,~N.~J.; Pelli,~D.~G. Deep learning—Using machine learning to study biological vision. \emph{Journal of vision} \textbf{2018}, \emph{18}, 2--2\relax
\mciteBstWouldAddEndPuncttrue
\mciteSetBstMidEndSepPunct{\mcitedefaultmidpunct}
{\mcitedefaultendpunct}{\mcitedefaultseppunct}\relax
\EndOfBibitem
\bibitem[Yadav \latin{et~al.}(2022)Yadav, Mollaei, Cao, Wang, and Farimani]{yadav2022prediction}
Yadav,~P.; Mollaei,~P.; Cao,~Z.; Wang,~Y.; Farimani,~A.~B. Prediction of GPCR activity using machine learning. \emph{Computational and Structural Biotechnology Journal} \textbf{2022}, \emph{20}, 2564--2573\relax
\mciteBstWouldAddEndPuncttrue
\mciteSetBstMidEndSepPunct{\mcitedefaultmidpunct}
{\mcitedefaultendpunct}{\mcitedefaultseppunct}\relax
\EndOfBibitem
\bibitem[Guntuboina \latin{et~al.}(2023)Guntuboina, Das, Mollaei, Kim, and Barati~Farimani]{guntuboina2023peptidebert}
Guntuboina,~C.; Das,~A.; Mollaei,~P.; Kim,~S.; Barati~Farimani,~A. PeptideBERT: A language model based on transformers for peptide property prediction. \emph{The Journal of Physical Chemistry Letters} \textbf{2023}, \emph{14}, 10427--10434\relax
\mciteBstWouldAddEndPuncttrue
\mciteSetBstMidEndSepPunct{\mcitedefaultmidpunct}
{\mcitedefaultendpunct}{\mcitedefaultseppunct}\relax
\EndOfBibitem
\bibitem[Mollaei and Barati~Farimani(2023)Mollaei, and Barati~Farimani]{mollaei2023activity}
Mollaei,~P.; Barati~Farimani,~A. Activity Map and Transition Pathways of G Protein-Coupled Receptor Revealed by Machine Learning. \emph{Journal of Chemical Information and Modeling} \textbf{2023}, \emph{63}, 2296--2304\relax
\mciteBstWouldAddEndPuncttrue
\mciteSetBstMidEndSepPunct{\mcitedefaultmidpunct}
{\mcitedefaultendpunct}{\mcitedefaultseppunct}\relax
\EndOfBibitem
\bibitem[Kim \latin{et~al.}(2024)Kim, Mollaei, Antony, Magar, and Barati~Farimani]{kim2024gpcr}
Kim,~S.; Mollaei,~P.; Antony,~A.; Magar,~R.; Barati~Farimani,~A. GPCR-BERT: Interpreting Sequential Design of G Protein-Coupled Receptors Using Protein Language Models. \emph{Journal of Chemical Information and Modeling} \textbf{2024}, \relax
\mciteBstWouldAddEndPunctfalse
\mciteSetBstMidEndSepPunct{\mcitedefaultmidpunct}
{}{\mcitedefaultseppunct}\relax
\EndOfBibitem
\bibitem[Baldi and Brunak(2001)Baldi, and Brunak]{baldi2001bioinformatics}
Baldi,~P.; Brunak,~S. \emph{Bioinformatics: the machine learning approach}; MIT press, 2001\relax
\mciteBstWouldAddEndPuncttrue
\mciteSetBstMidEndSepPunct{\mcitedefaultmidpunct}
{\mcitedefaultendpunct}{\mcitedefaultseppunct}\relax
\EndOfBibitem
\bibitem[Mollaei and Barati~Farimani(2023)Mollaei, and Barati~Farimani]{mollaei2023unveiling}
Mollaei,~P.; Barati~Farimani,~A. Unveiling Switching Function of Amino Acids in Proteins Using a Machine Learning Approach. \emph{Journal of Chemical Theory and Computation} \textbf{2023}, \emph{19}, 8472--8480\relax
\mciteBstWouldAddEndPuncttrue
\mciteSetBstMidEndSepPunct{\mcitedefaultmidpunct}
{\mcitedefaultendpunct}{\mcitedefaultseppunct}\relax
\EndOfBibitem
\bibitem[Mollaei \latin{et~al.}(2024)Mollaei, Sadasivam, Guntuboina, and Barati~Farimani]{mollaei2024idp}
Mollaei,~P.; Sadasivam,~D.; Guntuboina,~C.; Barati~Farimani,~A. IDP-Bert: Predicting Properties of Intrinsically Disordered Proteins Using Large Language Models. \emph{The Journal of Physical Chemistry B} \textbf{2024}, \emph{128}, 12030--12037\relax
\mciteBstWouldAddEndPuncttrue
\mciteSetBstMidEndSepPunct{\mcitedefaultmidpunct}
{\mcitedefaultendpunct}{\mcitedefaultseppunct}\relax
\EndOfBibitem
\bibitem[Camacho \latin{et~al.}(2018)Camacho, Collins, Powers, Costello, and Collins]{camacho2018next}
Camacho,~D.~M.; Collins,~K.~M.; Powers,~R.~K.; Costello,~J.~C.; Collins,~J.~J. Next-generation machine learning for biological networks. \emph{Cell} \textbf{2018}, \emph{173}, 1581--1592\relax
\mciteBstWouldAddEndPuncttrue
\mciteSetBstMidEndSepPunct{\mcitedefaultmidpunct}
{\mcitedefaultendpunct}{\mcitedefaultseppunct}\relax
\EndOfBibitem
\bibitem[Badrinarayanan \latin{et~al.}(2024)Badrinarayanan, Guntuboina, Mollaei, and Barati~Farimani]{badrinarayanan2024multi}
Badrinarayanan,~S.; Guntuboina,~C.; Mollaei,~P.; Barati~Farimani,~A. Multi-peptide: multimodality leveraged language-graph learning of peptide properties. \emph{Journal of Chemical Information and Modeling} \textbf{2024}, \emph{65}, 83--91\relax
\mciteBstWouldAddEndPuncttrue
\mciteSetBstMidEndSepPunct{\mcitedefaultmidpunct}
{\mcitedefaultendpunct}{\mcitedefaultseppunct}\relax
\EndOfBibitem
\bibitem[Kung(2014)]{kung2014kernel}
Kung,~S.~Y. \emph{Kernel methods and machine learning}; Cambridge University Press, 2014\relax
\mciteBstWouldAddEndPuncttrue
\mciteSetBstMidEndSepPunct{\mcitedefaultmidpunct}
{\mcitedefaultendpunct}{\mcitedefaultseppunct}\relax
\EndOfBibitem
\bibitem[Gehler \latin{et~al.}(2009)Gehler, Sch{\"o}lkopf, Camps-Valls, and Bruzzone]{gehler2009introduction}
Gehler,~P.~V.; Sch{\"o}lkopf,~B.; Camps-Valls,~G.; Bruzzone,~L. An introduction to kernel learning algorithms. \emph{Kernel methods for remote sensing data analysis} \textbf{2009}, 25--48\relax
\mciteBstWouldAddEndPuncttrue
\mciteSetBstMidEndSepPunct{\mcitedefaultmidpunct}
{\mcitedefaultendpunct}{\mcitedefaultseppunct}\relax
\EndOfBibitem
\bibitem[Abdi and Williams(2010)Abdi, and Williams]{abdi2010principal}
Abdi,~H.; Williams,~L.~J. Principal component analysis. \emph{Wiley interdisciplinary reviews: computational statistics} \textbf{2010}, \emph{2}, 433--459\relax
\mciteBstWouldAddEndPuncttrue
\mciteSetBstMidEndSepPunct{\mcitedefaultmidpunct}
{\mcitedefaultendpunct}{\mcitedefaultseppunct}\relax
\EndOfBibitem
\bibitem[Ma{\'c}kiewicz and Ratajczak(1993)Ma{\'c}kiewicz, and Ratajczak]{mackiewicz1993principal}
Ma{\'c}kiewicz,~A.; Ratajczak,~W. Principal components analysis (PCA). \emph{Computers \& Geosciences} \textbf{1993}, \emph{19}, 303--342\relax
\mciteBstWouldAddEndPuncttrue
\mciteSetBstMidEndSepPunct{\mcitedefaultmidpunct}
{\mcitedefaultendpunct}{\mcitedefaultseppunct}\relax
\EndOfBibitem
\bibitem[Doerr \latin{et~al.}(2021)Doerr, Majewski, P{\'e}rez, Kramer, Clementi, Noe, Giorgino, and De~Fabritiis]{doerr2021torchmd}
Doerr,~S.; Majewski,~M.; P{\'e}rez,~A.; Kramer,~A.; Clementi,~C.; Noe,~F.; Giorgino,~T.; De~Fabritiis,~G. TorchMD: A deep learning framework for molecular simulations. \emph{Journal of chemical theory and computation} \textbf{2021}, \emph{17}, 2355--2363\relax
\mciteBstWouldAddEndPuncttrue
\mciteSetBstMidEndSepPunct{\mcitedefaultmidpunct}
{\mcitedefaultendpunct}{\mcitedefaultseppunct}\relax
\EndOfBibitem
\bibitem[Kohlhoff \latin{et~al.}(2014)Kohlhoff, Shukla, Lawrenz, Bowman, Konerding, Belov, Altman, and Pande]{kohlhoff2014cloud}
Kohlhoff,~K.~J.; Shukla,~D.; Lawrenz,~M.; Bowman,~G.~R.; Konerding,~D.~E.; Belov,~D.; Altman,~R.~B.; Pande,~V.~S. Cloud-based simulations on Google Exacycle reveal ligand modulation of GPCR activation pathways. \emph{Nature chemistry} \textbf{2014}, \emph{6}, 15--21\relax
\mciteBstWouldAddEndPuncttrue
\mciteSetBstMidEndSepPunct{\mcitedefaultmidpunct}
{\mcitedefaultendpunct}{\mcitedefaultseppunct}\relax
\EndOfBibitem
\bibitem[Jorgensen \latin{et~al.}(1983)Jorgensen, Chandrasekhar, Madura, Impey, and Klein]{jorgensen1983comparison}
Jorgensen,~W.~L.; Chandrasekhar,~J.; Madura,~J.~D.; Impey,~R.~W.; Klein,~M.~L. Comparison of simple potential functions for simulating liquid water. \emph{The Journal of chemical physics} \textbf{1983}, \emph{79}, 926--935\relax
\mciteBstWouldAddEndPuncttrue
\mciteSetBstMidEndSepPunct{\mcitedefaultmidpunct}
{\mcitedefaultendpunct}{\mcitedefaultseppunct}\relax
\EndOfBibitem
\bibitem[Hellerstein \latin{et~al.}(2012)Hellerstein, Kohlhoff, and Konerding]{hellerstein2012science}
Hellerstein,~J.~L.; Kohlhoff,~K.~J.; Konerding,~D.~E. Science in the cloud: accelerating discovery in the 21st century. \emph{IEEE Internet Computing} \textbf{2012}, \emph{16}, 64--68\relax
\mciteBstWouldAddEndPuncttrue
\mciteSetBstMidEndSepPunct{\mcitedefaultmidpunct}
{\mcitedefaultendpunct}{\mcitedefaultseppunct}\relax
\EndOfBibitem
\bibitem[Hess \latin{et~al.}(2008)Hess, Kutzner, Van Der~Spoel, and Lindahl]{hess2008gromacs}
Hess,~B.; Kutzner,~C.; Van Der~Spoel,~D.; Lindahl,~E. GROMACS 4: algorithms for highly efficient, load-balanced, and scalable molecular simulation. \emph{Journal of chemical theory and computation} \textbf{2008}, \emph{4}, 435--447\relax
\mciteBstWouldAddEndPuncttrue
\mciteSetBstMidEndSepPunct{\mcitedefaultmidpunct}
{\mcitedefaultendpunct}{\mcitedefaultseppunct}\relax
\EndOfBibitem
\bibitem[Majewski \latin{et~al.}(2023)Majewski, P{\'e}rez, Th{\"o}lke, Doerr, Charron, Giorgino, Husic, Clementi, No{\'e}, and De~Fabritiis]{majewski2023machine}
Majewski,~M.; P{\'e}rez,~A.; Th{\"o}lke,~P.; Doerr,~S.; Charron,~N.~E.; Giorgino,~T.; Husic,~B.~E.; Clementi,~C.; No{\'e},~F.; De~Fabritiis,~G. Machine learning coarse-grained potentials of protein thermodynamics. \emph{Nature Communications} \textbf{2023}, \emph{14}, 5739\relax
\mciteBstWouldAddEndPuncttrue
\mciteSetBstMidEndSepPunct{\mcitedefaultmidpunct}
{\mcitedefaultendpunct}{\mcitedefaultseppunct}\relax
\EndOfBibitem
\bibitem[Kubelka \latin{et~al.}(2004)Kubelka, Hofrichter, and Eaton]{kubelka2004protein}
Kubelka,~J.; Hofrichter,~J.; Eaton,~W.~A. The protein folding ‘speed limit’. \emph{Current opinion in structural biology} \textbf{2004}, \emph{14}, 76--88\relax
\mciteBstWouldAddEndPuncttrue
\mciteSetBstMidEndSepPunct{\mcitedefaultmidpunct}
{\mcitedefaultendpunct}{\mcitedefaultseppunct}\relax
\EndOfBibitem
\bibitem[Lindorff-Larsen \latin{et~al.}(2011)Lindorff-Larsen, Piana, Dror, and Shaw]{lindorff2011fast}
Lindorff-Larsen,~K.; Piana,~S.; Dror,~R.~O.; Shaw,~D.~E. How fast-folding proteins fold. \emph{Science} \textbf{2011}, \emph{334}, 517--520\relax
\mciteBstWouldAddEndPuncttrue
\mciteSetBstMidEndSepPunct{\mcitedefaultmidpunct}
{\mcitedefaultendpunct}{\mcitedefaultseppunct}\relax
\EndOfBibitem
\bibitem[Harvey \latin{et~al.}(2009)Harvey, Giupponi, and Fabritiis]{harvey2009acemd}
Harvey,~M.~J.; Giupponi,~G.; Fabritiis,~G.~D. ACEMD: accelerating biomolecular dynamics in the microsecond time scale. \emph{Journal of chemical theory and computation} \textbf{2009}, \emph{5}, 1632--1639\relax
\mciteBstWouldAddEndPuncttrue
\mciteSetBstMidEndSepPunct{\mcitedefaultmidpunct}
{\mcitedefaultendpunct}{\mcitedefaultseppunct}\relax
\EndOfBibitem
\bibitem[Buch \latin{et~al.}(2010)Buch, Harvey, Giorgino, Anderson, and De~Fabritiis]{buch2010high}
Buch,~I.; Harvey,~M.~J.; Giorgino,~T.; Anderson,~D.~P.; De~Fabritiis,~G. High-throughput all-atom molecular dynamics simulations using distributed computing. \emph{Journal of chemical information and modeling} \textbf{2010}, \emph{50}, 397--403\relax
\mciteBstWouldAddEndPuncttrue
\mciteSetBstMidEndSepPunct{\mcitedefaultmidpunct}
{\mcitedefaultendpunct}{\mcitedefaultseppunct}\relax
\EndOfBibitem
\bibitem[Piana \latin{et~al.}(2011)Piana, Lindorff-Larsen, and Shaw]{piana2011robust}
Piana,~S.; Lindorff-Larsen,~K.; Shaw,~D.~E. How robust are protein folding simulations with respect to force field parameterization? \emph{Biophysical journal} \textbf{2011}, \emph{100}, L47--L49\relax
\mciteBstWouldAddEndPuncttrue
\mciteSetBstMidEndSepPunct{\mcitedefaultmidpunct}
{\mcitedefaultendpunct}{\mcitedefaultseppunct}\relax
\EndOfBibitem
\bibitem[Rosenbaum \latin{et~al.}(2009)Rosenbaum, Rasmussen, and Kobilka]{rosenbaum2009structure}
Rosenbaum,~D.~M.; Rasmussen,~S.~G.; Kobilka,~B.~K. The structure and function of G-protein-coupled receptors. \emph{Nature} \textbf{2009}, \emph{459}, 356--363\relax
\mciteBstWouldAddEndPuncttrue
\mciteSetBstMidEndSepPunct{\mcitedefaultmidpunct}
{\mcitedefaultendpunct}{\mcitedefaultseppunct}\relax
\EndOfBibitem
\bibitem[Katritch \latin{et~al.}(2013)Katritch, Cherezov, and Stevens]{katritch2013structure}
Katritch,~V.; Cherezov,~V.; Stevens,~R.~C. Structure-function of the G protein--coupled receptor superfamily. \emph{Annual review of pharmacology and toxicology} \textbf{2013}, \emph{53}, 531--556\relax
\mciteBstWouldAddEndPuncttrue
\mciteSetBstMidEndSepPunct{\mcitedefaultmidpunct}
{\mcitedefaultendpunct}{\mcitedefaultseppunct}\relax
\EndOfBibitem
\bibitem[Sch{\"o}neberg \latin{et~al.}(1999)Sch{\"o}neberg, Schultz, and Gudermann]{schoneberg1999structural}
Sch{\"o}neberg,~T.; Schultz,~G.; Gudermann,~T. Structural basis of G protein-coupled receptor function. \emph{Molecular and cellular endocrinology} \textbf{1999}, \emph{151}, 181--193\relax
\mciteBstWouldAddEndPuncttrue
\mciteSetBstMidEndSepPunct{\mcitedefaultmidpunct}
{\mcitedefaultendpunct}{\mcitedefaultseppunct}\relax
\EndOfBibitem
\bibitem[Wang \latin{et~al.}(2018)Wang, Qiao, and Li]{wang2018new}
Wang,~W.; Qiao,~Y.; Li,~Z. New insights into modes of GPCR activation. \emph{Trends in pharmacological sciences} \textbf{2018}, \emph{39}, 367--386\relax
\mciteBstWouldAddEndPuncttrue
\mciteSetBstMidEndSepPunct{\mcitedefaultmidpunct}
{\mcitedefaultendpunct}{\mcitedefaultseppunct}\relax
\EndOfBibitem
\bibitem[Heng \latin{et~al.}(2013)Heng, Aubel, and Fussenegger]{heng2013overview}
Heng,~B.~C.; Aubel,~D.; Fussenegger,~M. An overview of the diverse roles of G-protein coupled receptors (GPCRs) in the pathophysiology of various human diseases. \emph{Biotechnology advances} \textbf{2013}, \emph{31}, 1676--1694\relax
\mciteBstWouldAddEndPuncttrue
\mciteSetBstMidEndSepPunct{\mcitedefaultmidpunct}
{\mcitedefaultendpunct}{\mcitedefaultseppunct}\relax
\EndOfBibitem
\bibitem[Hilger \latin{et~al.}(2018)Hilger, Masureel, and Kobilka]{hilger2018structure}
Hilger,~D.; Masureel,~M.; Kobilka,~B.~K. Structure and dynamics of GPCR signaling complexes. \emph{Nature structural \& molecular biology} \textbf{2018}, \emph{25}, 4--12\relax
\mciteBstWouldAddEndPuncttrue
\mciteSetBstMidEndSepPunct{\mcitedefaultmidpunct}
{\mcitedefaultendpunct}{\mcitedefaultseppunct}\relax
\EndOfBibitem
\bibitem[Simpson \latin{et~al.}(2011)Simpson, Wall, Blaney, and Reynolds]{simpson2011modeling}
Simpson,~L.~M.; Wall,~I.~D.; Blaney,~F.~E.; Reynolds,~C.~A. Modeling GPCR active state conformations: The $\beta$2-adrenergic receptor. \emph{Proteins: Structure, Function, and Bioinformatics} \textbf{2011}, \emph{79}, 1441--1457\relax
\mciteBstWouldAddEndPuncttrue
\mciteSetBstMidEndSepPunct{\mcitedefaultmidpunct}
{\mcitedefaultendpunct}{\mcitedefaultseppunct}\relax
\EndOfBibitem
\bibitem[Latorraca \latin{et~al.}(2017)Latorraca, Venkatakrishnan, and Dror]{latorraca2017gpcr}
Latorraca,~N.~R.; Venkatakrishnan,~A.; Dror,~R.~O. GPCR dynamics: structures in motion. \emph{Chemical reviews} \textbf{2017}, \emph{117}, 139--155\relax
\mciteBstWouldAddEndPuncttrue
\mciteSetBstMidEndSepPunct{\mcitedefaultmidpunct}
{\mcitedefaultendpunct}{\mcitedefaultseppunct}\relax
\EndOfBibitem
\bibitem[Ali \latin{et~al.}(2020)Ali, Naveed, Gordon, Majeed, Saeed, Ogbuke, Atif, Zubair, and Changxing]{ali2020beta}
Ali,~D.~C.; Naveed,~M.; Gordon,~A.; Majeed,~F.; Saeed,~M.; Ogbuke,~M.~I.; Atif,~M.; Zubair,~H.~M.; Changxing,~L. $\beta$-Adrenergic receptor, an essential target in cardiovascular diseases. \emph{Heart failure reviews} \textbf{2020}, \emph{25}, 343--354\relax
\mciteBstWouldAddEndPuncttrue
\mciteSetBstMidEndSepPunct{\mcitedefaultmidpunct}
{\mcitedefaultendpunct}{\mcitedefaultseppunct}\relax
\EndOfBibitem
\bibitem[Yang \latin{et~al.}(2021)Yang, Yu, Wu, and Wang]{yang2021role}
Yang,~A.; Yu,~G.; Wu,~Y.; Wang,~H. Role of $\beta$2-adrenergic receptors in chronic obstructive pulmonary disease. \emph{Life Sciences} \textbf{2021}, \emph{265}, 118864\relax
\mciteBstWouldAddEndPuncttrue
\mciteSetBstMidEndSepPunct{\mcitedefaultmidpunct}
{\mcitedefaultendpunct}{\mcitedefaultseppunct}\relax
\EndOfBibitem
\bibitem[Cherezov \latin{et~al.}(2007)Cherezov, Rosenbaum, Hanson, Rasmussen, Thian, Kobilka, Choi, Kuhn, Weis, Kobilka, \latin{et~al.} others]{cherezov2007high}
Cherezov,~V.; Rosenbaum,~D.~M.; Hanson,~M.~A.; Rasmussen,~S.~G.; Thian,~F.~S.; Kobilka,~T.~S.; Choi,~H.-J.; Kuhn,~P.; Weis,~W.~I.; Kobilka,~B.~K.; others High-resolution crystal structure of an engineered human $\beta$2-adrenergic G protein--coupled receptor. \emph{science} \textbf{2007}, \emph{318}, 1258--1265\relax
\mciteBstWouldAddEndPuncttrue
\mciteSetBstMidEndSepPunct{\mcitedefaultmidpunct}
{\mcitedefaultendpunct}{\mcitedefaultseppunct}\relax
\EndOfBibitem
\bibitem[Rasmussen \latin{et~al.}(2011)Rasmussen, Choi, Fung, Pardon, Casarosa, Chae, DeVree, Rosenbaum, Thian, Kobilka, \latin{et~al.} others]{rasmussen2011structure}
Rasmussen,~S.~G.; Choi,~H.-J.; Fung,~J.~J.; Pardon,~E.; Casarosa,~P.; Chae,~P.~S.; DeVree,~B.~T.; Rosenbaum,~D.~M.; Thian,~F.~S.; Kobilka,~T.~S.; others Structure of a nanobody-stabilized active state of the $\beta$2 adrenoceptor. \emph{Nature} \textbf{2011}, \emph{469}, 175--180\relax
\mciteBstWouldAddEndPuncttrue
\mciteSetBstMidEndSepPunct{\mcitedefaultmidpunct}
{\mcitedefaultendpunct}{\mcitedefaultseppunct}\relax
\EndOfBibitem
\bibitem[McGibbon \latin{et~al.}(2015)McGibbon, Beauchamp, Harrigan, Klein, Swails, Hern{\'a}ndez, Schwantes, Wang, Lane, and Pande]{mcgibbon2015mdtraj}
McGibbon,~R.~T.; Beauchamp,~K.~A.; Harrigan,~M.~P.; Klein,~C.; Swails,~J.~M.; Hern{\'a}ndez,~C.~X.; Schwantes,~C.~R.; Wang,~L.-P.; Lane,~T.~J.; Pande,~V.~S. MDTraj: a modern open library for the analysis of molecular dynamics trajectories. \emph{Biophysical journal} \textbf{2015}, \emph{109}, 1528--1532\relax
\mciteBstWouldAddEndPuncttrue
\mciteSetBstMidEndSepPunct{\mcitedefaultmidpunct}
{\mcitedefaultendpunct}{\mcitedefaultseppunct}\relax
\EndOfBibitem
\end{mcitethebibliography}
\end{document}


\maketitle

\section{1: Data Preprocessing}

\textbf {I. Reference protein structure:} \\
To simulate protein dynamics, water and/or ions may be added to the protein’s environment. To isolate only the protein structure, the surrounding environment must be purified. For the unfolded structure of small proteins, a chain of residues can also be constructed using molecular visualization tools such as Visual Molecular Dynamics (VMD)\cite{humphrey1996vmd}.
In our study, we built protein B, NTL9, Trp-Cage and Chignolin through VMD and chose the purified inactive structure of the $\beta_2$ adrenergic receptor (PDB 2RH1)\cite{cherezov2007high} as the reference structures.

\begin{figure}[t!]
     \centering
     \includegraphics[width=\linewidth]{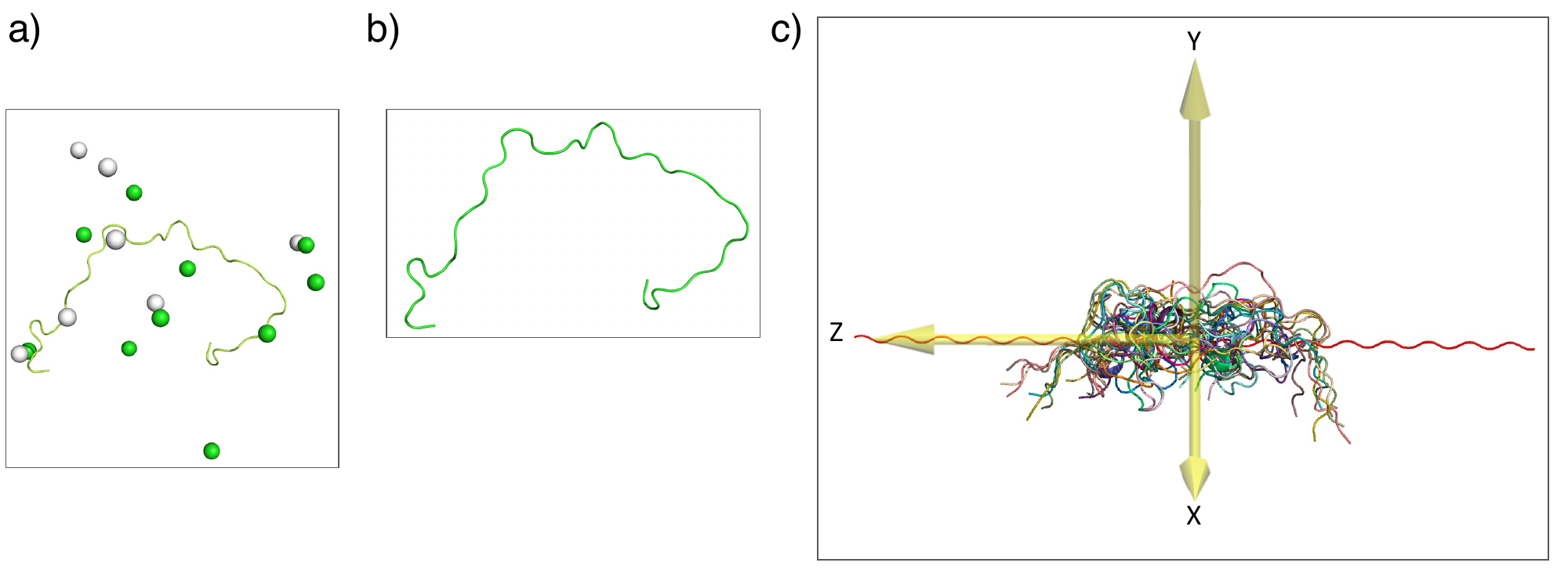}
     \caption {(a) Original protein structure with additional elements from MD simulations, (b) isolated protein structure after removing extra elements, and (c) aligned trajectory structures with respect to the oriented reference protein.}
     \label{figs:ntl9prot}
 \end{figure}

To build the protein structure, follow these steps:

\begin{enumerate}

\item  From components of a simulation (Figure \ref{figs:ntl9prot}a), select only the protein sequence (Figure \ref{figs:ntl9prot}b).

\item Build the reference protein structure using VMD (Main \textrightarrow Extensions \textrightarrow Modeling \textrightarrow Molefacture \textrightarrow Build \textrightarrow Protein Builder \textrightarrow Add the sequence from part 1\textrightarrow Select Straight \textrightarrow  Build)

\item Save the PDB file of the reference protein. Later, you will align trajectory structures to this one (Figure \ref{figs:ntl9prot}c).

\item (Optional): Open the PDB file in text format. If the order of atoms in the generated PDB file does not match that of the trajectories, or if the residue indices are not sequentially ordered (e.g., the N atom of residue \#2 is located among the atoms of residue \#1), open the PDB file in PyMOL \cite{delano2002pymol} and save it. Go to PyMOL \textrightarrow File \textrightarrow Export Molecule (choose pdb format). Later, repeat this process for the trajectories.
\end{enumerate}

\noindent \textbf {II. Optional: Draw a sphere at the origin and shift the geometric center of the reference protein to it}
\begin{enumerate}
\item Load the reference protein in VMD. Use the "move\_to\_origin.tcl" command (see GitHub) in the TKConsole or specify it in the VMD startup command.
\end{enumerate}

\noindent \textbf {III. Align principal axes of the reference protein with the x, y and z axes}.
\begin{enumerate}
\item Download orient.tar.gz (here\href{http://www.ks.uiuc.edu/Research/vmd/script_library/scripts/orient/orient.tar.gz}) and la101psx.tar.gz (here\href{http://www.ks.uiuc.edu/Research/vmd/script_library/scripts/orient/la101psx.tar.gz}). The latter is a linear algebra package by Hume Integration Software; more information can be found on this website: http://www.hume.com/la/.
\item Unpack and load the packages and orient the protein using the "alignment\_package.tcl" (see GitHub)
\item Save the oriented reference protein.
\end{enumerate}

\noindent \textbf {IV. Prepare simulation trajectories}
\begin{enumerate}
\item If the simulation trajectories contain extra elements, save only the essential protein residues using the "Selected atoms" section in VMD (for example, resid 2 to 9). Ensure that the saved residues are compatible with the reference protein.
\end{enumerate}

\noindent \textbf {V. Align simulation trajectories to the reference protein}
\begin{enumerate}
\item Load both the reference protein and the simulation trajectories into VMD. Ensure that both files display the same number of atoms on the VMD main screen.
\item Align the entire simulation trajectory with the reference protein. Extensions \textrightarrow Analysis \textrightarrow RMSD Trajectory Tools. Choose "Top" for the "reference mol" (ensure that the 'T' on the VMD main screen is associated with the reference protein) and select "Backbone" for the "Selection Modifiers".
\item Save the aligned simulation trajectories. They will be used as input for the Kernel model.
\end{enumerate}

\clearpage
\section{2: Protein Representation Using Atomic Coordinates}
\begin{figure}[t!]
 \centering
 \includegraphics[width=0.75\linewidth]{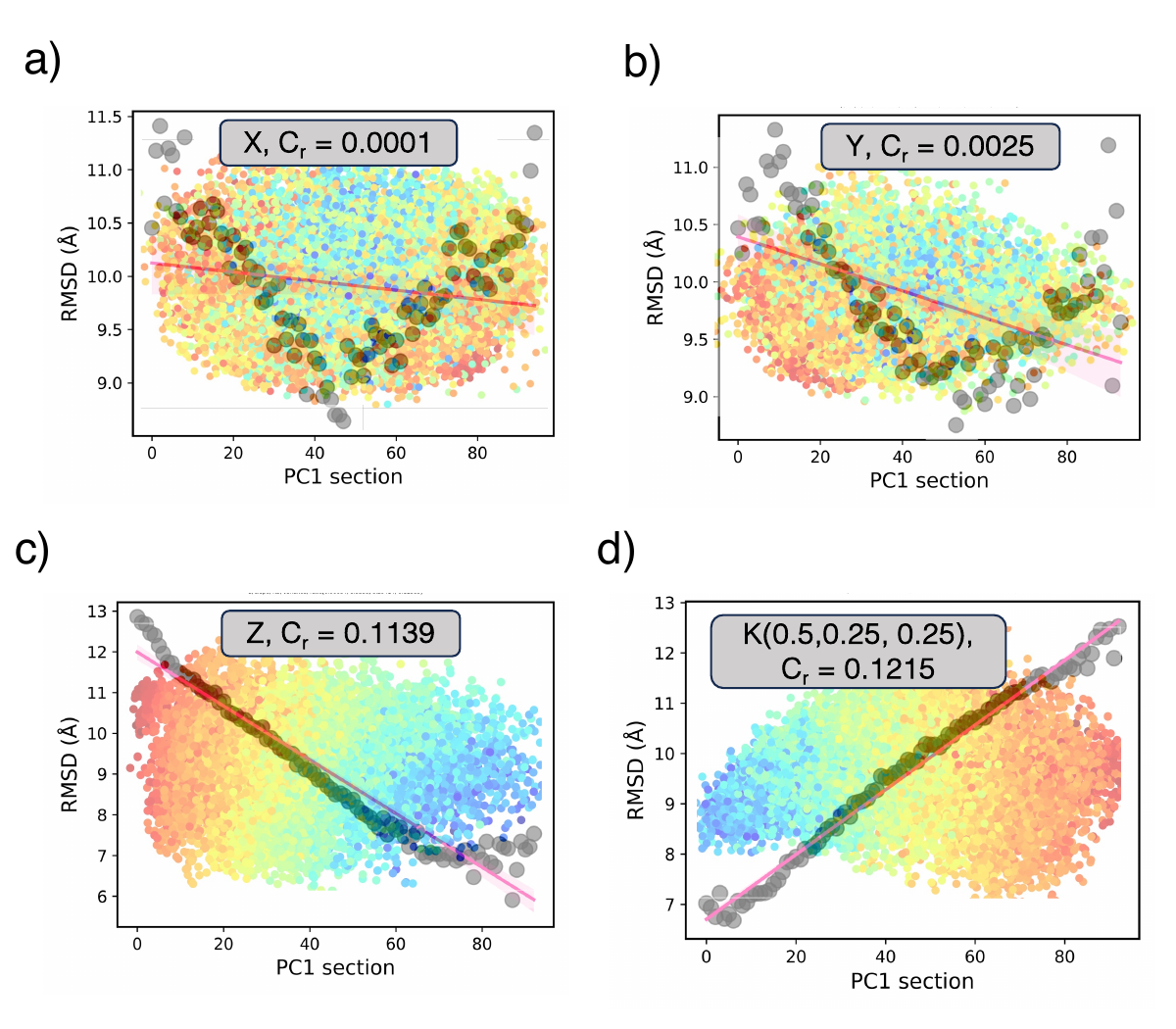}
 \caption{Protein representations using raw atomic coordinates: (a) x, (b) y, and (c) z in the Trp-Cage protein (PDB: 2JOF). (d) The optimal representation obtained using the Kernel-PCA model.}
 \label{fig:xyz}
\end{figure}

The preprocessed trajectories using only x, y and z coordinates of atoms may or may not yield an appropriate representation associated with the protein property, depending on the dynamic complexity of the protein. Figures \ref{fig:xyz} illustrate the RMSD projected onto the optimal representations generated solely from the x (a), y (b), and z (c) coordinates of atoms in the Trp-Cage protein (PDB: 2JOF). This analysis shows that, in terms of the Correlation ratio, the z-coordinate of atoms provides a better representation compared to the x and y coordinates. However, Figure \ref{fig:xyz}d demonstrates that applying Kernel-PCA results in a representation with a higher Correlation ratio, offering a stronger association with RMSD. Thus, to achieve a generalizable representation that effectively captures key reaction coordinates that influence global properties, we utilized the Kernel-PCA model.

\clearpage
\section{3: Correlation Ratios Correspond to K($\lambda_1$, $\lambda_2$, $\lambda_3$)}

\begin{figure}[t!]
     \centering
     \includegraphics[width=0.7\linewidth]{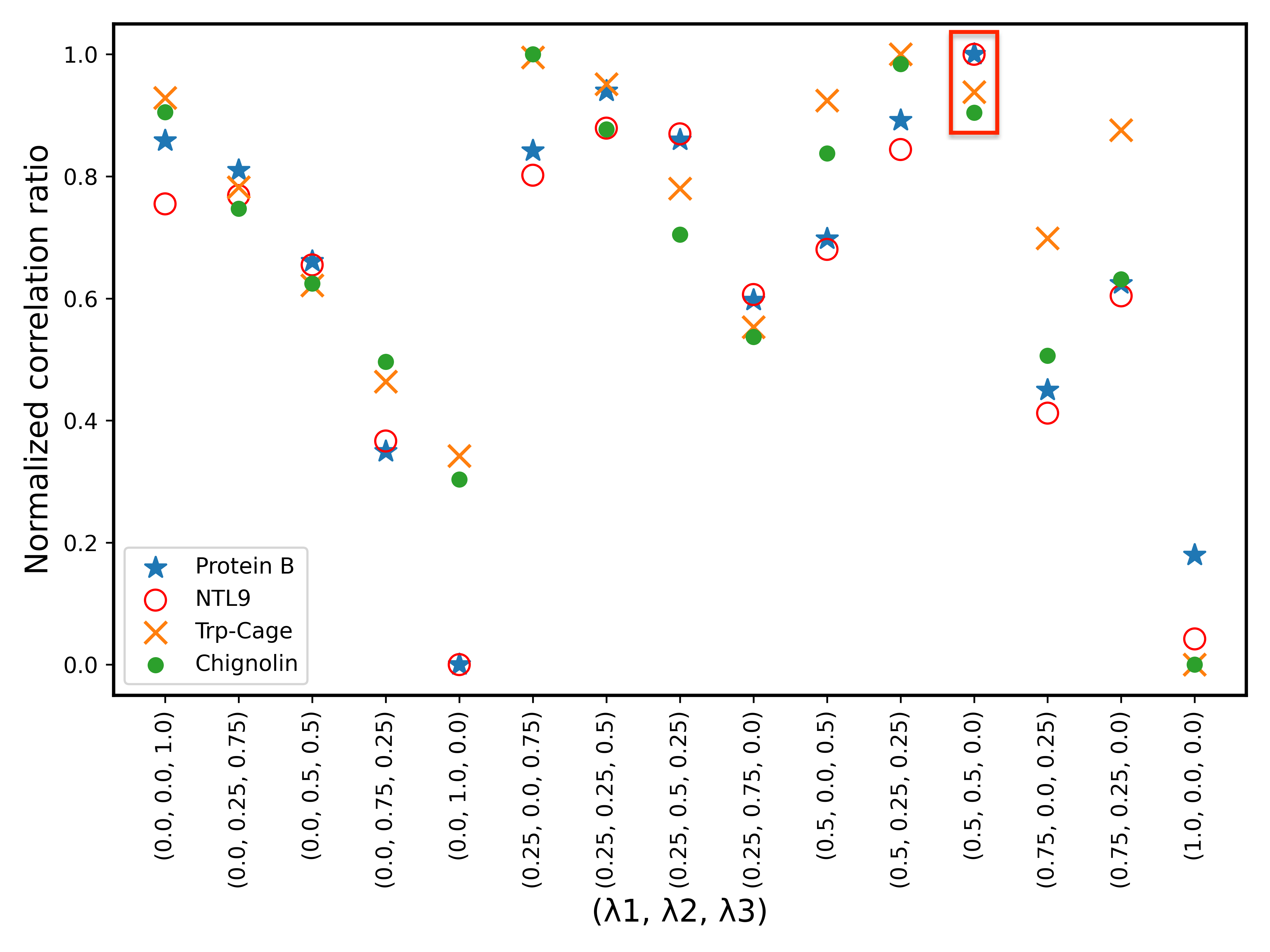}
     \caption{Normalized Correlation ratio obtained from different combinations of ($\lambda_1$, $\lambda_2$, $\lambda_3$) for Protein B, NTL9, Trp-Cage, and Chignolin.}
     \label{fig:lam}
 \end{figure}

Table~S1 reports the best and worst ($\lambda_1$, $\lambda_2$, $\lambda_3$) corresponding to the highest and lowest Correlation ratio. Figure\ref{fig:lam} represents normalized Correlation ratios vs ($\lambda_1$, $\lambda_2$, $\lambda_3$) for Protein B, NTL9, Trp-Cage, and Chignolin. Although a specific combination ($\lambda_1$, $\lambda_2$, $\lambda_3$) could provide the highest Correlation ratio for each protein, the combination (0.5, 0.5, 0), highlighted on the figure, can generate the representation with a high Correlation ratio across different proteins.

\begin{table}[t!]
\caption{The $\lambda_1$, $\lambda_2$, $\lambda_3$ values corresponding to the highest and lowest Correlation ratios.} 
\includegraphics[width=0.4\linewidth]{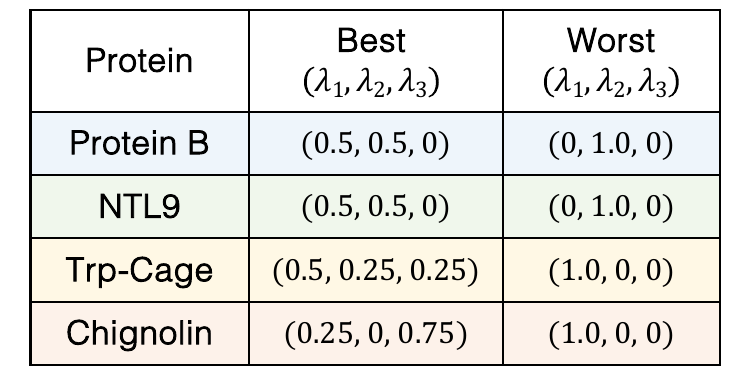}
\label{tab:table 2} 
\end{table}

\section{4: Top Reaction Coordinates in $\beta_2$ Adrenergic Receptor}

\begin{figure}[t!]
 \centering
 \includegraphics[width=\linewidth]{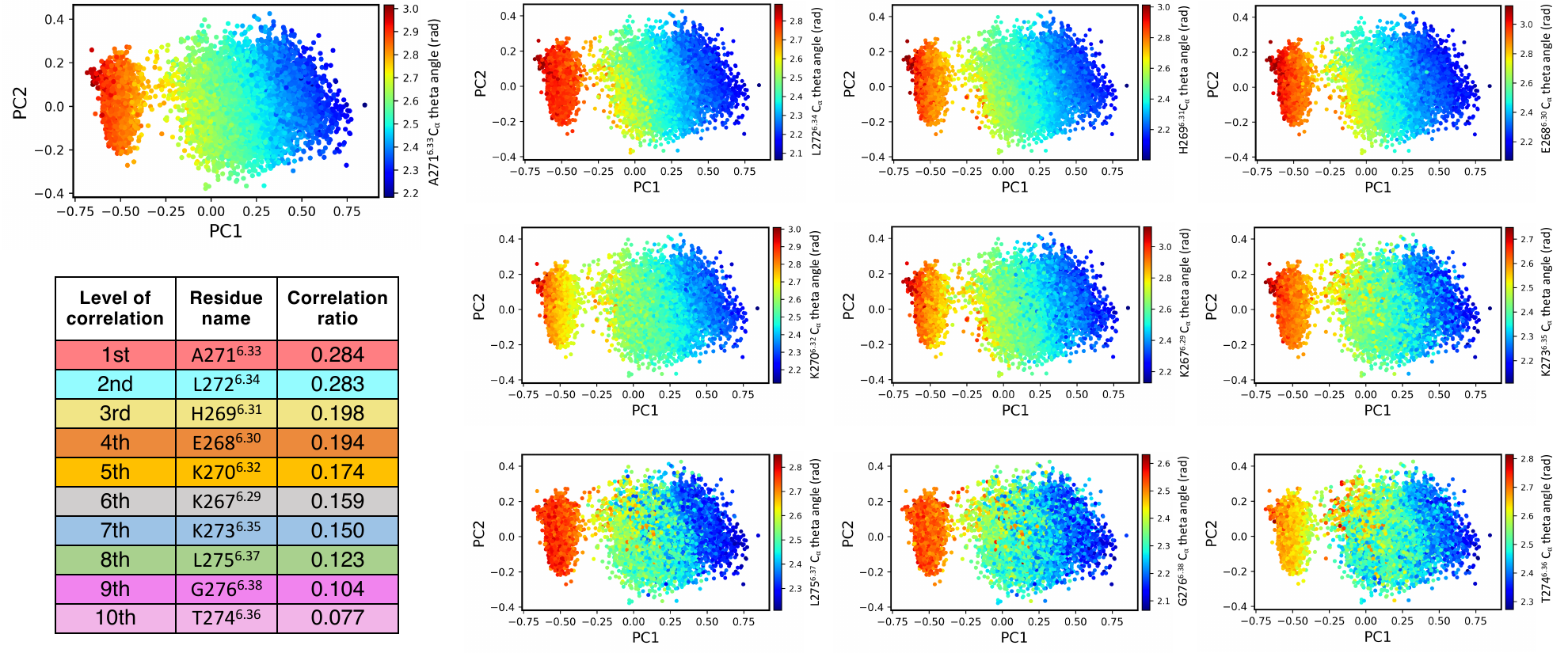}
 \caption{Top ten reaction coordinates of the $\beta_2$ Adrenergic Receptor, with each coordinate showing the dynamics of the theta angle in the $CA$ atom of each residue, projected onto the optimal representation.}
 \label{fig:top10}
\end{figure}

Figure\ref{fig:top10} illustrates the top ten reaction coordinates in $\beta_2$AR, representing the theta angles of the CA atoms in the residues most correlated to the receptor activation process. As described in Figure 3, all the top ten residues are located in the intracellular part of TM6, where significant conformational changes occur during the activation process. Comparison of Figure S3 with Figure 3.b reveals a strong connection between dynamics of the top ten residues, both among themselves and with the activation state of the receptor.

\clearpage
\section{5: Relationships Between Top Reaction Coordinates in Protein B, Trp-Cage, and Chignolin Proteins}

\begin{figure}[t!]
     \centering
     \includegraphics[width=\linewidth]{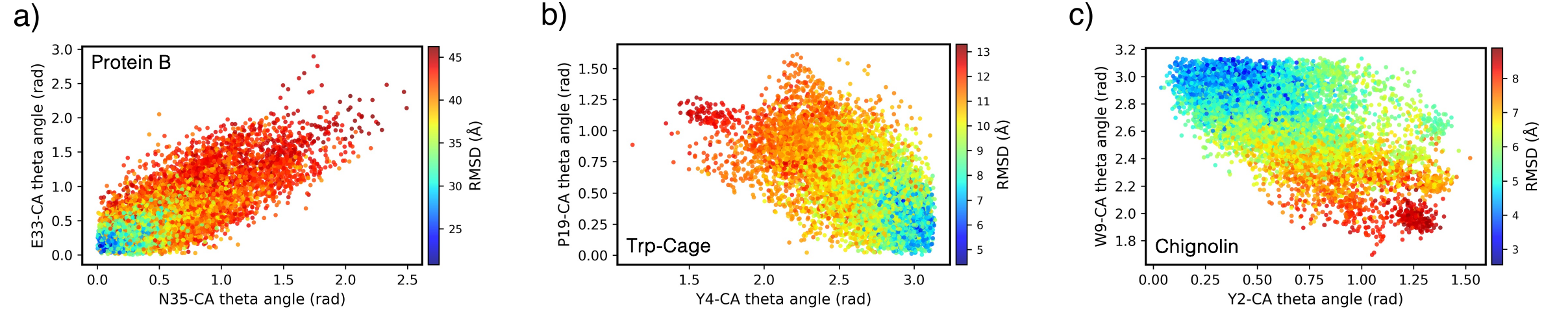}
     \caption{Top two reaction coordinates for (a) Protein B, (b) Trp-Cage, and (c) Chignolin proteins corresponding to the theta angles of CA atoms in the top two amino acids correlated with the folding state of each protein.}
     \label{fig:2topatoms_3prot}
 \end{figure}

Figure\ref{fig:2topatoms_3prot} shows the relationships between the top two reaction coordinates corresponding to theta angles in two most associated residues with the folding states in protein B (a), Trp-Cage (b), and Chignolin (c) proteins. This information provides insight into the correlation between dynamics and conformations of these residues and their relationship with the overall protein property and function.

\bibliography{reference}